# The Missing Knowledge Layer in AI

## A Framework for Stable Human–AI Reasoning

Rikard Rosenbacke*, Carl Rosenbacke[1], Victor Rosenbacke[1,2], Martin McKee[3]

[1]*Faculty of Medicine, Lund University, Sweden*
[2]*Department of Economics, Lund University School of Economics and Management, Sweden*
[3]*Department of Health Services Research and Policy, London School of Hygiene & Tropical Medicine, UK*
*Corresponding Author: rikard@rosenbacke.com

Version 2.0 - April 2026

## Abstract

Large language models are increasingly integrated into everyday decision-making in areas such as healthcare, law, finance, engineering, and government. Yet they share a crucial limitation that is easy to miss: they sound confident even when their internal reasoning has drifted. A fluent answer can hide uncertainty, speculation, or confusion, and small changes in phrasing can produce entirely different conclusions. This makes LLMs powerful assistants, but unreliable partners in high-stakes decisions.

Humans have similar weaknesses. We also mistake fluency for truth. When a model replies smoothly, we naturally trust it, even when both the model and the human are drifting together. This paper is the first in a five-paper research series examining how stable human–AI reasoning can be achieved. The series proposes a two-layer response: Parts II–IV develop simple human-side mechanisms - uncertainty cues, conflict surfacing, and auditable reasoning traces - that stabilise interpretation at decision points. Part V develops the model-side counterpart: an inference-time Epistemic Control Loop (ECL) that detects instability, contradiction, and drift during generation and modulates decoding accordingly. In the same way that humans pause when something "feels off", the ECL gives models a basic form of that self-monitoring.

When combined, the human-side stabilisation layer (Layer 1) and the model-side Epistemic Control Loop (Layer 2) form the missing operational substrate for governance: they raise signal-to-noise at the point of use, before enforcement is applied. Only once interaction is stabilised - with reliance, uncertainty, and drift made legible on both sides - can Layer 3 capability governance (who gets access, what is logged, what triggers escalation, and how incidents are investigated) operate precisely rather than bluntly. This ordering principle - stabilisation before capability governance - also aligns with emerging compliance expectations under the EU AI Act, ISO/IEC 42001, and clinical governance, because it makes the reasoning process traceable under real conditions of use and not only in the abstract terms found in many policy documents.

The central message is simple: **Fluency is not reliability. And without a structure that stabilises both humans and models, AI cannot be trusted or governed where it matters most.** This paper explains why, and what we can do about it.

## Structure of the Five-Paper Research Series

This paper is part of a five-paper research series examining how stable human–AI reasoning can be achieved by regulating not only model capabilities but also the human–AI relationship itself.

**Part I,** *The Missing Knowledge Layer in AI: A Framework for Stable Human–AI Reasoning*[1] This paper identifies a system-wide gap and sets out why it matters, motivating the need for a dedicated knowledge layer that can govern human–AI reasoning, considering both human and model contributions, rather than, as is usual, focusing exclusively on model-level capabilities.

**Part II,** *Beyond "Hallucinations": A Framework for Stable Human–AI Reasoning*[2] diagnoses the problem by analysing **why** human users systematically mistake linguistic fluency for epistemic understanding, and how large language models, trained on natural language, optimise for coherence and plausibility rather than factual correctness, leading to predictable forms of reasoning collapse.

**Part III,** *Governing Reflective Human–AI Collaboration: A Framework for Epistemic Scaffolding and Traceable Reasoning*[3] proposes **what** must be governed: the human–AI relationship itself, introducing mechanisms for epistemic scaffolding, traceability, and alignment with emerging regulatory frameworks (e.g., the EU AI Act, WHO digital health standards, ISO guidelines).

**Part IV,** *From Consumption to Reflection: Designing Human–AI Relations for Stable Reasoning*[4] shows **how** this relationship can be implemented at the interface level, treating interaction design as the primary unit of analysis for stabilising reasoning during use.

**Part V,** *Epistemic Control Loops in Large Language Models: An Architectural Proposal for Machine-Side Regulation*[5] provides a technical proposal for inference-time regulation within language models, including internal epistemic signals, control loops, actuation mechanisms, and memory. This component is currently under development and will be released in a subsequent preprint.

This paper serves as the entry point to the series. It motivates the overall framework, situates the problem at a system-wide level, and explains why downstream solutions, human-side governance, interface design, and machine-side regulation require a dedicated knowledge layer to be effective. Readers interested in the cognitive mechanisms underlying human error may proceed to Part II, while those focused on governance or technical implementation may consult Parts III–V.

# 1. Epistemic Collapse: A Shared Human–AI Failure Mode

When we try to make sense of the world, we rarely pause to examine the nature of our own thinking. We draw on memory, inference, intuition, and learned patterns, yet, in conscious experience these different modes of cognition all present themselves as a single thing: *knowledge*. A conclusion feels like a fact. Intuition is a direct perception. A familiar story feels like reality itself.

This is where a seemingly abstract philosophical distinction becomes practically important. Ontology concerns what exists, what is actually out there in the world. Epistemology concerns what we *believe* about what exists, and how well-justified those beliefs are. In principle, the distinction is straightforward. In practice, however, humans often treat epistemic representations as if they were ontological facts. This well supported by decades of cognitive science and behavioural economics. Humans systematically under-track uncertainty and conflate inference with fact, making this a structural feature of cognition rather than an occasional error. The two collapse into each other with surprising ease.

This slippage is widespread even among experts. Models, heuristics, rules of thumb, and provisional abstractions are experienced as direct descriptions of the world rather than as fallible maps of a complex terrain. Scientific disciplines formally distinguish between data, inference, and conjecture, yet in day-to-day reasoning, these boundaries frequently blur. Confidence, coherence, or disciplinary consensus is implicitly taken to imply truth. The map–territory distinction remains conceptually clear but operationally fragile. This paper, Part I of the series, is the synthesis and ordering paper.

In Parts II–IV of this series,[2–4] which contain the detailed cognitive, interface, governance, and model-side proposals, we will show that epistemic collapse is not a sporadic event but a structural feature of human cognition. People rarely experience their beliefs *as* beliefs. They often experience them as facts, even when the underlying reasoning is tentative or weakly grounded. Across domains, such as clinical decision-making, economics, expert forecasting, and policy analysis, humans struggle to track whether a conclusion is retrieved from memory, inferred from evidence, or constructed through intuitive pattern completion. Epistemic status is largely invisible; subjective coherence is overwhelming.

Large language models (LLMs) inherit this cognitive vulnerability in amplified form. Trained on human-generated text, where epistemic markers are sparse, implicit, or inconsistent, LLMs internalise patterns in which memory, inference, speculation, and narrative extrapolation are all expressed with the same surface fluency. Here, a caveat is necessary. Inevitably, given their recent development, the empirical basis for this form of epistemic collapse is somewhat weaker. Thus, while emerging work reports correlations between hallucination risk or reasoning fragility and signals such as divergence across samples, perturbation sensitivity, or hidden-state dynamics, these signals remain probabilistic, model-specific, and task-dependent, and so do not yet constitute a stable or universal epistemic indicator. Hence, references in this paper to epistemic state or instability signals in LLMs should be understood as *hypothesised regulatory affordances*, not as fully validated diagnostics.

Crucially, unlike humans, LLMs lack any metacognitive regulation that could compensate for this collapse. They do not reveal whether a token is produced through recall, deductive pattern completion, abductive extrapolation, or unsupported guesswork. Autoregressive decoding proceeds smoothly even when internal activations begin to drift or conflict.

This structural absence explains why LLM failures are patterned rather than random:

- **Hallucinations** reflect unrecognised transitions from grounded retrieval to speculative interpolation.
- **Overconfidence** arises because all outputs share an undifferentiated surface confidence level.
- **False coherence** appears when internally unstable trajectories still yield linguistically smooth text.

Humans mitigate epistemic collapse imperfectly through hesitation, second-order reasoning, doubt, and social accountability. These mechanisms fail under cognitive load or time pressure, but they exist. Transformers have no analogous layers. Once a trajectory is initiated, generation continues without epistemic gating: the model has no internal signal indicating that it has shifted into a fragile or drifting mode of reasoning.

When humans interact with fluent models, these vulnerabilities compound. Fluency is mistaken for epistemic grounding. Agreement between a user's intuition and the model's output is misread as mutual confirmation, even when both are wrong. In Part II, we term this pattern *false confirmation*. This joint failure mode is particularly dangerous in clinical, legal, financial, and policy contexts, where errors accumulate silently and may carry systemic consequences.

The implication is straightforward: improving factual accuracy, scaling model capacity, or adding more training data does not resolve the underlying problem. As long as neither humans nor models explicitly represent epistemic status, both will continue to express their outputs as though they were ontologically grounded. This is why hallucinations persist even in models that are otherwise impressively capable.

This analysis motivates the ordering principle that is developed in the remaining parts of this series. Parts II–IV show that epistemic collapse can be mitigated, to some extent, through reflective interface mechanisms and governance structures that help users recognise uncertainty, differentiate recall from inference, and resist false confirmation. But enhancing human reasoning alone is insufficient if the model continues to drift internally under paraphrase, multi-turn pressure, or long-horizon task structure. Part V, therefore, proposes an architectural counterpart: inference-time mechanisms within LLMs that monitor and regulate epistemic instability during generation. Only by stabilising both sides of the relation can we prevent epistemological uncertainty from collapsing into an illusion of ontological fact.

**The Ordering Principle: Stabilisation Before Capability Governance**
A central claim of this series is that the dominant mode of governance failure is not a lack of rules but a lack of signal. Capability governance and deployment control - who gets access, what gets logged, what triggers intervention, and how incidents are investigated - become blunt when they are applied downstream of epistemically degraded interaction signals. For example, if a deployed model produces harmful medical or legal advice, post-hoc logs often cannot distinguish whether this was deliberate misuse, ordinary user confusion, interface-induced overreliance, or model-side drift under paraphrase - yet governance is expected to assign responsibility, trigger enforcement, and update access policy on that basis. Much of today's AI safety discourse therefore oscillates between model-internal alignment ("the brain") and external governance ("the law"), while missing the operational substrate in between: the interaction-time stabilisation and telemetry layer that determines what signals institutions, auditors, and monitoring systems actually receive.

When we argue for stabilisation prior to capability governance, we do not mean to suggest that governance can be postponed until epistemic signals are clean, or that existing regulatory practice is naive about uncertainty. In reality, governance regimes routinely operate under noisy, partial, and contested information, using proxies, thresholds, audits, and discretion to manage risk. Our claim is narrower. When the signals produced during human–AI interaction are epistemically degraded, governance necessarily becomes coarse: attribution is difficult, enforcement is blunt, and post hoc reconstruction carries a high evidentiary burden. Stabilisation does not replace governance, nor does it precede it in time. Rather, it increases the *resolution* at which governance can operate, allowing institutions to move from broad, precautionary controls toward more precise, proportionate, and justifiable intervention.

Recent frontier governance work in 2026 increasingly points to the same structural bottleneck: governance is being asked to operate on signals that are unstable, incomplete, or can only be reconstructed after the fact. The *International AI Safety Report 2026* emphasises the limits of evaluation, the difficulty of monitoring deployed systems, and the challenge of governing autonomy under uncertainty and competition.[6] Amodei's *The Adolescence of Technology* frames the problem similarly: frontier capability is becoming too consequential for informal norms and ad hoc safeguards, and must be governed as infrastructure rather than as a conventional product.[7] NATO's strategic assessment makes the signal problem explicit in institutional language, warning that "*Knowledge is a strategic currency*," and that "*Advanced AI systems may fragment evidence… shifting the burden of proof… to choosing which AI-mediated system has the authority to define what constitutes valid evidence*."[8]

The three-layer architecture developed across this series is designed as a direct response to this signal degradation. Layer 1 stabilises the human side of interaction by making epistemic posture, reliance, and responsibility legible at the moment decisions are formed (Part II-IV). Layer 2 stabilises the model side by turning runtime instability into measurable signals and a controllable state during inference (Part V). Only once Layers 1–2 raise the signal-to-noise ratio does Layer 3 capability governance and deployment control become precise rather than politically brittle. The claim is not that stabilisation replaces governance, but that it makes governance operationally feasible by ensuring that enforcement, evaluation, and accountability operate on clean signals rather than fluent noise.

The following section, therefore, begins with the human side of the problem, asking how users can be supported in engaging in reflective reasoning even when interacting with fluent yet epistemically unstable systems.

## 2. The Human Side (Our Trilogy)

The central claim in Parts II–IV is that humans drift epistemically when interacting with fluent systems because language collapses uncertainty. When an LLM expresses a guess, an inference, and a memory with the same surface confidence, the user cannot reliably distinguish which kind of cognitive act they are responding to. This collapse reshapes how humans interpret information: uncertainty becomes invisible, and intuition fills the gap. The result is not a failure of knowledge but a failure of epistemic awareness.

Human–AI interfaces amplify this vulnerability. As will be shown in Part II (*Beyond "Hallucinations"*), fluency is cognitively seductive: coherent output is mistaken for grounded output, and agreement is mistaken for evidence. This invokes Kahneman's dual-process theory, in which System 1 denotes fast, intuitive processing, while System 2 refers to slower, reflective reasoning.[9] The user's System 1 accepts what is fluent before System 2 has a chance to interrogate it. Part III (*Governing Reflective Human–AI Collaboration*) shows that common interface patterns, chat windows, rapid turn-taking, and conversational framing reinforce this automatic absorption. They encourage consumption rather than reflection, making epistemic drift the default trajectory of interaction.

A trilogy of papers (Part II-IV) then argues that humans require deliberate scaffolding to counteract these tendencies. Part III introduces the idea that reflective reasoning can be architected: epistemic prompts, reasoning cues, conflict framing, and structured counterfactuals can slow the user's cognition enough to reintroduce epistemic discrimination. Part IV (*From Consumption to Reflection*) elaborates on this by presenting a relational model: the interface is not merely a channel but a partner in thought. When well designed, it guides the user into a reflective mode in which uncertainty is visible, disagreement is expected, and the epistemic status of a conclusion is part of its representation. This forms the human-side epistemic control loop, an interaction pattern that makes the user aware not only of what they conclude but also of how they reason.

Stabilising the human side alone is not sufficient. Users may be supported into a more reflective posture, yet the model they interact with can still drift internally, shifting from retrieval to speculation, or from stable reasoning to fragile extrapolation, without signalling that change. This asymmetry limits the effectiveness of any human-side safeguard. The trilogy, therefore, points to the need for a complementary model-side architecture: if humans require scaffolds to regulate their epistemic dynamics, then models require mechanisms to regulate theirs as well.

The present paper addresses this gap by proposing the Epistemic Control Loop, the machine-side analogue that enables stable, mutually accountable reasoning across the human–AI system.

## 3. The Missing Layer: A Model-Side Epistemic Regulator

If the unit of analysis is the human–AI relationship, then stabilising one side of the system is not enough. The trilogy will argue that humans drift because language collapses uncertainty, and that reflective scaffolds can restore part of the lost epistemic structure. But even the most reflective user cannot compensate for a model that lacks an internal mechanism to track its reasoning. Human-side improvement addresses only half of the relational loop. The other half, the model's epistemic dynamics, remains ungoverned. This is the second step in the ordering principle introduced in Section 1: stabilisation before capability governance.

Large language models generate text through a fluent autoregressive process that monitors no internal epistemic distinctions at all. A token produced through recall, one made through deductive pattern completion, one produced through abductive extrapolation, and one created through unsupported guesswork appear identical on the surface. The model neither represents nor signals the difference, and it does not adjust its behaviour when the underlying mode of reasoning shifts. From the outside, all outputs look equally confident because the model has no mechanism that would allow them to differ.

Internally, however, these modes of reasoning are not identical. Hidden states may exhibit different patterns of stability, dispersion, and trajectory drift depending on whether the model is retrieving learned associations or improvising a continuation under uncertainty. These internal signatures appear to yield potentially informative signals, though their clarity and consistency vary across architectures. Some of these may be measurable: divergence across stochastic samples, disagreement across latent reasoning paths, changes in covariance structure, and perturbation sensitivity can all correlate with impending hallucination or epistemic instability. The problem is not that the signals are absent, but that the model lacks an architectural component to interpret them.

This absence becomes critical in complex high-stakes tasks. Legal analysis, medical deliberation, financial decisions, and multi-step strategic planning all impose epistemic requirements that go far beyond surface coherence. They demand continuity across steps, discrimination between stable and fragile inferences, and the ability to detect when an internal trajectory is drifting. Yet autoregressive decoding treats every next token as equally valid, provided it is locally fluent. A model may be internally unstable while appearing perfectly confident. Fluency masks drift.

The problem becomes more acute in emerging agentic systems. When models operate autonomously across many sequential reasoning steps - planning actions, calling tools, observing results, and updating strategies - small epistemic uncertainties can accumulate and amplify. Early speculative inferences may propagate into later decisions, gradually shifting the system's trajectory without any explicit failure signal. In such settings, the risk is not a single hallucinated token but rather progressive epistemic drift, in which an initially weak assumption shapes an expanding chain of reasoning. As language models move from assistants responding to isolated prompts toward agents operating across extended decision horizons, inference-time stabilisation becomes increasingly important for maintaining coherent and reliable reasoning over time.

In such settings, the relational failure becomes structural. The user cannot see the model's epistemic instability; the model has no mechanism to see it either. Errors do not announce themselves. They accumulate silently inside fluent language. Human-side scaffolds can help users interrogate their own reasoning, but they cannot provide the model with the introspection it lacks. As a result, the human–AI system remains unstable even when the human alone is stabilised.

What is missing is a model-side regulator: a mechanism that identifies the model's current mode of reasoning, detects internal divergence before errors appear, adapts generation when uncertainty rises, and maintains commitments across steps. In other words, the model needs the epistemic counterpart to the human reflective scaffolds developed in Parts II–IV. Without such a regulator, large language models will continue to operate with unexamined epistemic assumptions, and users will continue to overread fluency as a form of grounding.

The preceding analysis makes the gap clear: the human–AI system cannot become epistemically stable unless the model acquires the capacity to monitor and regulate its own reasoning. Transformers were never designed for this function; they provide fluency without epistemic structure. What is required is an additional layer, not a new architecture, but a regulatory process that governs how the model reasons rather than what it outputs.

## 4. The Proposal: The Epistemic Control Loop (High-Level)

This section introduces a proposal for the regulatory layer. The Epistemic Control Loop (ECL)[5] provides the model-side counterpart to the reflective scaffolds for humans that we will report in Parts II–IV. Its purpose is straightforward: to give the model an operational sense of how it is reasoning, when its internal dynamics are becoming unstable, and when it should slow down, verify, or maintain continuity across steps. The ECL does not replace the Transformer backbone or alter its parameters. The term "control loop" is used in a functional rather than formal sense; it does not imply provable stability or convergence guarantees. It adds epistemic functions that the architecture lacks, functions that become necessary when models participate in shared reasoning with humans.

The ECL is not a new architecture. It is a **regulatory layer** that sits on top of existing Transformers and governs how they generate text. It consists of four conceptual components:

### i) Epistemic State, The model must know how it is reasoning

Humans distinguish (imperfectly) between memory, inference, and guessing. LLMs do not: all modes of generation share the same surface fluency. Yet internally, models appear to leave informative patterns in the form of hidden-state configurations that correlate with stable retrieval, others with more fragile extrapolation. We use the term epistemic state, drawing on the late-19th-century introduction of the adjective epistemic ("pertaining to knowledge") from Greek *epistēmē*, to denote a compact, operational description of these internal activation patterns as they relate to different reasoning modes. This does not imply correctness or metacognition; it simply characterises how the model is generating a token at a given step.

A growing empirical literature shows that LLM hidden states are not epistemically uniform. Probing studies demonstrate that models encode semantically and functionally distinct information in separable subspaces.[10] Work on hallucination prediction and uncertainty modelling shows that internal divergence, associated with semantic entropy, variance across stochastic samples, or perturbation sensitivity, tends to increase when models move from grounded recall to unstable extrapolation.[10] Self-consistency analyses suggest that agreement across sampled reasoning paths can indicate the stability of inference.[11] These insights suggest that internal signatures of reasoning modes are measurable even when the model does not explicitly represent them.

Operationally, we propose that an epistemic state can be approximated using indicators such as: (i) stability or divergence of hidden-state trajectories; (ii) agreement across stochastic decoding passes; (iii) alignment or conflict among attention heads; and (iv) shifts in activation covariance. For example, when answering well-known facts, trajectories tend to converge and attention patterns remain focused; under ambiguity, stochastic samples and internal activations typically diverge more widely. The epistemic state abstracts these observable differences into a heuristic, functional classification usable by the Epistemic Control Loop. By this, we do not mean that the model recognises or understands its own reasoning. Rather, the model does not "know how it is reasoning"; instead, the control loop treats certain recurring internal patterns *as epistemically meaningful* to regulate behaviour under uncertainty. We therefore emphasise that epistemic state is a theoretical construct grounded in emerging empirical evidence, not a claim that reasoning modes are fully understood or introspectively accessible. Current results are sufficient, however, to justify using epistemic state as a practical control variable for asking the structural question: **"What type of generative behaviour is occurring right now?"**

These signals should be understood as comparative rather than absolute. They do not provide a general measure of knowledge quality, but indicate relative changes in generative behaviour across nearby states. Even partial or noisy distinctions may therefore be sufficient for regulation.

### ii) Instability Signal, Detect divergence before errors surface

Humans sense conflict before they speak. Models do so internally as well. Independent reasoning paths diverge when uncertainty rises; stochastic samples disagree; hidden-state trajectories drift. Models do not respond to these signals, but we believe they exist. We do, however, concede that our understanding of this concept remains tentative. Divergence may be model- or task-specific, and the correlation with hallucination may be weak or inconsistent. Consequently, the instability signal should be treated as a promising hypothesis rather than a validated diagnostic. We also recognise the need to consider alternative mechanisms that may explain the observed divergence, such as prompt sensitivity, in which small lexical changes produce large activation shifts unrelated to uncertainty, or gaps in the training data, in which the model encounters inputs far from its learned distribution. Accordingly, we present the instability signal as an evolving construct grounded in early evidence, with its reliability, scope, and architectural generality to be tested in future work that we propose later in this series of papers. Nonetheless, while recognising these limitations, we see the instability signal as the collection of these cues into a measurable indicator of epistemic stress: **"Is the model entering a region of unstable reasoning?"**

The instability signal should therefore be understood as a heuristic control variable rather than a definitive diagnostic of error, supporting adaptive responses under uncertainty rather than certifying failure.

### iii) Regulated Actuation, Slow down when uncertain

When humans notice doubt, they pause, consider alternatives, or check their steps. Models do not. Autoregressive generation proceeds at full speed even when internal states destabilise. Regulated actuation introduces adaptive behaviour: reducing sampling entropy, triggering verification, revisiting earlier steps, or allocating more compute. The goal is not censorship but self-regulation: **"Given my current epistemic state, should I think more before speaking?"**

### iv) Epistemic Memory, Maintain continuity across steps

Humans do not abandon stable beliefs without justification. Transformers do. They can contradict themselves within a few tokens. Epistemic memory stores minimal information about prior low-stress commitments and uses it to detect unjustified reversals. It is not a world model; it is a consistency check: **"Does this new conclusion conflict with something I previously said with high confidence?"**

### Relation to *Attention Is All You Need*

The ECL does not modify the Transformer; it builds alongside it, drawing on the properties introduced in *Attention Is All You Need*,[12] the architecture underlying all modern LLMs. Self-attention and autoregressive decoding already generate internal signals that can support epistemic regulation when appropriately interpreted. Four features of the Transformer make this possible:

**Rich internal representations:** Self-attention produces high-dimensional hidden states that encode relational patterns. Empirical probing studies show that certain activation signatures correlate with stability, retrieval, or drift, providing raw signals for an epistemic regulator to interpret.

**Parallel attention patterns:** Different attention heads compute distinct weighted representations in parallel. When these heads diverge, the resulting differences provide a natural signal of potential inconsistencies or uncertainty across the model's latent reasoning paths.

**Autoregressive decoding:** Transformers generate text token by token. This process can be modulated in real time by adjusting sampling parameters or invoking additional reasoning steps when internal uncertainty signals rise without altering the underlying model.

**Limited cross-turn coherence:** Transformers maintain context only within their attention window and do not track commitments over time. External memory mechanisms supply this missing function, enabling consistency checks across longer interactions.

None of these properties was initially designed for epistemic governance. The historical objective of language modelling was to maximise fluency and predictive accuracy, not to distinguish recall from speculation or detect internal drift. As LLMs have moved into high-stakes domains, such as clinical reasoning, legal analysis, and financial decision-making, the absence of explicit epistemic control has become more salient. The Epistemic Control Loop addresses this gap without modifying the Transformer backbone, adding a regulatory layer that interprets and manages the internal signals the architecture already produces.

### A High-Level Framework, Not an Engineering Blueprint

The ECL should be understood as a conceptual synthesis rather than an engineering specification. Its components draw on established strands of work, some of which we have described above when discussing instability signals. They include chain-of-thought consistency methods that aim to improve reliability by seeking agreement across sampled reasoning paths.[11] However, they operate *externally* and do not track the model's internal epistemic dynamics. Hierarchical or planning-oriented models offer structured, multi-step reasoning but focus on task decomposition rather than on detecting internal drift.[13]

Several uncertainty-aware decoding techniques, including Bayesian ensembling (combining predictions from multiple models or multiple runs of the same model in a way that approximates Bayesian inference), semantic entropy (how much a model's generated interpretations diverge in meaning, used to estimate its uncertainty), and variance-based predictors, estimate output uncertainty.[10] but typically provide scalar confidence measures rather than a representation of the model's internal reasoning mode. Retrieval-augmented verification and verifier models, such as DeepMind's verifier frameworks, assess factual accuracy, but do not regulate how internal states evolve during generation. Approaches such as constitutional AI and self-supervised oversight introduce normative constraints on output behaviour, rather than on inference-time epistemic monitoring.[14] Similarly, interpretability-driven governance provides diagnostic tools but is not typically integrated into real-time generation.[15]

Our contribution lies in treating these techniques as parts of a coherent regulatory system rather than isolated methods. We do not present ourselves as machine-learning modellers, nor do we propose new training regimes or architectural modifications. Our focus is architectural and cognitive: to outline a model-side epistemic control loop that complements the human-side framework that we develop in Parts II–IV.

The more detailed technical pathways, including candidate implementations, design choices, and supporting references, are set out in Part V: *Epistemic Control Loops in LLMs*. Readers interested in how the components might be instantiated in practice can consult that paper. The present document remains non-technical by design; it sets out the governance logic and conceptual structure that motivate the approach, rather than the engineering needed to realise it.

## 5. Why This Matters (Economy, Governance, Safety)

Modern AI development rests on an implicit economic assumption: that scale will eventually produce stability. This assumption underlies the unprecedented global CAPEX commitments, estimated at US$ 5–7 trillion by 2030, directed toward larger models, larger clusters, and ever-expanding compute.[16,17] Yet stability is not guaranteed as a by-product of scale. It is an architectural property. Without explicit mechanisms that regulate how models reason and how humans interpret their outputs, capability growth amplifies the very fragilities institutions depend on AI to reduce.

The trilogy (Parts II–IV) will show that humans routinely misread linguistic fluency as epistemic grounding, that LLMs collapse uncertainty into surface coherence, and that interfaces often intensify this illusion. The human side, therefore, requires structured scaffolds, conflict cues, uncertainty signals, counterfactual prompts, and deliberate pauses to keep reasoning anchored and falsifiable. When these scaffolds are activated in high-stakes settings, the system produces an *Auditable Reasoning Trace (ART)*: a lightweight, timestamped record that links user intent, model output, reflective steps, conflicts surfaced, uncertainties considered, and the final judgement. An ART is not surveillance; it is a cognitive audit trail. It preserves the reasoning path behind a decision so that it can be reviewed, contested, verified, and improved. This directly satisfies the expectations embedded in the EU AI Act (transparency, traceability, meaningful human oversight), WHO digital standards (documented reasoning processes), and ISO 42001 (auditable risk assessment and control). The trilogy, therefore, provides a compliance-ready human layer: a method for stabilising user reasoning while producing documented, decision tracesthat can be reconstructed and are required in high-risk domains.

But human governance alone is insufficient. Institutions also need consistent behaviour from the model itself. Without a model-side regulator, small changes in phrasing or context can produce different answers; multi-step tasks drift across sessions; and errors accumulate silently in clinical, financial, or policy workflows. In this sense, the current LLM ecosystem resembles the early web before Google's PageRank: vast capability, but no stable interpretive layer to organise it. Before PageRank, search engines returned pages based on crude keyword matching. Small changes in wording could completely reorganise results. There was no mechanism that stabilised meaning, structure, or relevance across queries. PageRank did not add new content; it added a *governance layer* that made the web usable. It imposed coherence on an otherwise unstable information landscape. LLMs today sit in the same pre-PageRank state. They generate fluent answers, but lack the regulatory layer that stabilises reasoning across prompts, time, and context. Fluency accelerates adoption, but instability impedes deployment in environments where consistency and traceability are critical. What PageRank achieved for information retrieval, bringing order and stability to an otherwise chaotic system, epistemic regulation must now achieve for machine reasoning.

This is the structural case for the Epistemic Control Loop. The ECL does not replace scaling, fine-tuning, or world-model research. It provides the missing regulatory layer that ensures those capabilities behave predictably in institutional settings. By monitoring epistemic state, detecting internal conflict, modulating decoding under uncertainty, and maintaining cross-turn coherence, the model becomes a more stable partner in the human–AI relationship defined in the trilogy. The two layers, human scaffolding and model-side regulation, form the first integrated system capable of meeting both cognitive and regulatory requirements for high-stakes deployment. In practice, this means that capability governance can only become precise once interaction-time stabilisation has raised the signal-to-noise ratio in the human–AI relationship.

Although this paper is not an engineering blueprint, adopting epistemic regulation does not rely solely on normative commitment. Institutions already pay high hidden costs to compensate for model instability through human review, defensive prompting, conservative deployment, and legal risk buffering. Epistemic regulation shifts part of this burden from downstream governance and labour to interaction-time stabilisation. For frontier labs, it reduces reputational and liability risk as models enter institutional workflows; for enterprises, it lowers variance and improves auditability; for regulators, it supplies usable signals without intrusive oversight. In this sense, the Epistemic Control Loop addresses a governance bottleneck analogous to PageRank: stabilising trust and attribution rather than surface performance.

The economic implications follow directly. The projected productivity gains that justify global AI investment rely on sound reasoning in domains where errors incur financial, ethical, or societal costs. Enterprises cannot embed models in clinical pathways, legal reviews, financial risk analysis, procurement, or governmental decision-making processes unless the outputs are stable over time and auditable across cases. Regulators will not accept black-box reasoning. Boards cannot assume responsibility for decisions they cannot reconstruct. In short, neither institutional trust nor regulatory compliance can be achieved without explicit epistemic governance.

This three-layer approach, human scaffolding plus model-side ECL, creates the capability governance architecture that bridges capability and deployment. It satisfies regulatory expectations, supports organisational auditability, and allows enterprises to rely on models without incurring unmanageable epistemic risk. Without this architecture, the US$5 trillion investment cycle remains vulnerable to a structural gap: powerful models that are impressive in isolation but unreliable in the workflows where economic value is actually realised.

The aim of this framework is therefore straightforward. If AI is to serve as a reasoning partner in regulated, decision-heavy environments, both sides of the human–AI system must be epistemically stabilised. The trilogy provides the human-side layer; the ECL introduces the model-side equivalent. Together, they form a coherent cognitive infrastructure, arguably the missing meta-layer in the present AI stack, capable of supporting safe, traceable, institution-grade reasoning at scale.

## Implementation Pathways

The dual-layer epistemic governance architecture can be implemented at multiple levels. Frontier Labs can integrate ECL components directly into their model APIs, treating epistemic regulation as a first-class product feature alongside latency and cost optimisation. Enterprise platforms can build ECL functionality as middleware that works across multiple base models, analogous to how observability layers sit above compute infrastructure. Regulatory bodies can incorporate auditable reasoning traces and epistemic monitoring into compliance frameworks for high-risk AI applications.

The technical specifications in Part V are designed to be model-agnostic and API-compatible, allowing gradual adoption without requiring wholesale architectural changes. Organisations can begin with human-side scaffolds (trilogy) while pilot-testing model-side regulation (ECL), progressively strengthening the governance layer as both components mature.

## Conclusion

Human reasoning becomes unstable when interacting with fluent but epistemically flat models, as will be developed in Parts II–IV. Users over-trust coherence, underestimate uncertainty, and drift into false confirmation. To counter this, the human side of the relationship requires reflective scaffolds, conflict cues, counterfactual prompts, uncertainty signals, and the Auditable Reasoning Trace (ART). These mechanisms stabilise judgement at decision points and provide the traceability, oversight, and accountability increasingly expected in regulated settings.

Part V extends the same stabilisation logic to the model. Because transformers do not monitor how they form conclusions, they require an inference-time epistemic control loop: a lightweight regulator that detects epistemic stress, characterises reasoning mode, modulates decoding under uncertainty, and reduces contradiction and drift across steps. This is not a training innovation but an architectural addition, complementary to the human-side stabilisation rather than a substitute for it.

The core insight across all five papers is that the unit of analysis is neither the human nor the model in isolation, but the relationship between them. Reasoning failures emerge in the interaction: human overconfidence amplifies model drift, and model fluency reinforces human misinterpretation. Stability, therefore, requires a dual-stabilisation layer: Layer 1 on the human side and Layer 2 on the model side.

This leads to the ordering principle. Capability governance and deployment control - who gets access, what is logged, what triggers escalation, and how incidents are investigated - must often operate on signals generated during human–AI interaction. When those signals are epistemically degraded, attribution becomes uncertain, and enforcement risks becoming blunt or reliant on post hoc reconstruction.

When Layers 1–2 raise the signal-to-noise ratio in the interaction itself, Layer 3 capability governance can operate with far greater precision. Stabilised reasoning traces make it easier to distinguish misuse from misunderstanding, interface effects from model drift, and isolated errors from systemic failure. In this sense, interaction-time stabilisation does not replace governance; it strengthens the evidentiary signals on which governance relies.

Taken together, the series proposes an integrated epistemic governance architecture for human–AI collaboration. It provides the cognitive and technical stability required for deployment in clinical pathways, legal workflows, financial analysis, procurement, engineering design, and public administration. It also provides a concrete route to satisfy emerging documentation and oversight expectations (EU AI Act, WHO digital health standards, ISO/IEC 42001) through operational stabilisation rather than bureaucracy.

This is the foundation on which institution-grade human–AI reasoning can be built.